\pdfoutput=1
\documentclass[11pt]{article}

\usepackage[]{naacl2021}

\usepackage{times}
\usepackage{latexsym}

\usepackage[T1]{fontenc}
\usepackage[utf8]{inputenc}
\usepackage{url}
\usepackage{microtype}
\usepackage[pdftex]{graphicx}
\usepackage{multirow}
\usepackage{amsfonts}
\usepackage{amssymb}
\usepackage{amsmath}
\usepackage{booktabs}
\usepackage{xcolor}
\usepackage{bm}
\usepackage{numprint}
\usepackage{forest}
\usepackage{tikz-qtree}

\DeclareMathOperator*{\argmax}{arg\,max}

\title{Video-aided Unsupervised Grammar Induction }

\author{
Songyang Zhang$^1$\Thanks{~This work was done when Songyang Zhang was an intern at Tencent AI Lab.}, Linfeng Song$^2$, Lifeng Jin$^2$, Kun Xu$^2$, Dong Yu$^2$ \and Jiebo Luo$^1$ \\
$^1$University of Rochester, Rochester, NY, USA\\ 
\texttt{szhang83@ur.rochester.edu}, \texttt{jluo@cs.rochester.edu}\\
$^2$Tencent AI Lab, Bellevue, WA, USA \\ 
\texttt{\{lfsong,lifengjin,kxkunxu,dyu\}@tencent.com}\\ 
}

\begin{document}
\maketitle
\begin{abstract}
We investigate video-aided grammar induction, which learns a constituency parser from both unlabeled text and its corresponding video.
Existing methods of multi-modal grammar induction focus on learning syntactic grammars from text-image pairs, with promising results showing that the information from static images is useful in induction.
% ignoring the fact that nowadays an increasing number of texts in social media come with videos.
% Intuitively, videos are essential for recognizing verb and adverb phrases, which often carry important information.
However, videos provide even richer information, including not only static objects but also actions and state changes useful for inducing verb phrases. 
In this paper, we explore rich features (\textit{e.g.} action, object, scene, audio, face, OCR and speech) from videos, taking the recent Compound PCFG model \cite{kim2019compound} as the baseline. 
We further propose a Multi-Modal Compound PCFG model (MMC-PCFG) to effectively aggregate these rich features from different modalities.
Our proposed MMC-PCFG is trained end-to-end and outperforms each individual modality and previous state-of-the-art systems on three benchmarks, \textit{i.e.} DiDeMo, YouCook2 and MSRVTT, confirming the effectiveness of leveraging video information for unsupervised grammar induction.

\end{abstract}

\section{Introduction}

Constituency parsing is an important task in natural language processing, which
aims to capture syntactic information in sentences in the form of constituency parsing trees.
Many conventional approaches learn constituency parser from human-annotated datasets such as Penn Treebank~\cite{marcus-etal-1993-building}. 
However, annotating syntactic trees by human language experts is expensive and time-consuming, while the supervised approaches are limited to several major languages.
In addition, the treebanks for training these supervised parsers are small in size and restricted to the newswire domain, thus their performances tend to be worse when applying to other domains \cite{fried-etal-2019-cross}.
To address these issues, recent approaches \cite{shen2018ordered,jin2018unsupervised,drozdov2019unsupervised,kim2019compound} design unsupervised constituency parsers and grammar inducers, since they can be trained on large-scale unlabeled data.
In particular, there has been growing interests in exploiting visual information for unsupervised grammar induction because visual information can capture important knowledge required for language learning that is ignored by text
%textual information lacks 
\citep{Gleitman1990,pinker1987bootstrapping,Tomasello2003}. 
%because such data is popular in many social media platforms (e.g. Twitter and Facebook).
This task aims to learn a constituency parser from raw unlabeled text aided by its visual context.
% This task aims to learn a constituency parser from both unlabeled text and its visual context.
% It is based on the assumption that similar syntactic constituents are usually matched to similar visual concepts.

\begin{figure}[t]
\centering
\includegraphics[width=0.5\textwidth]{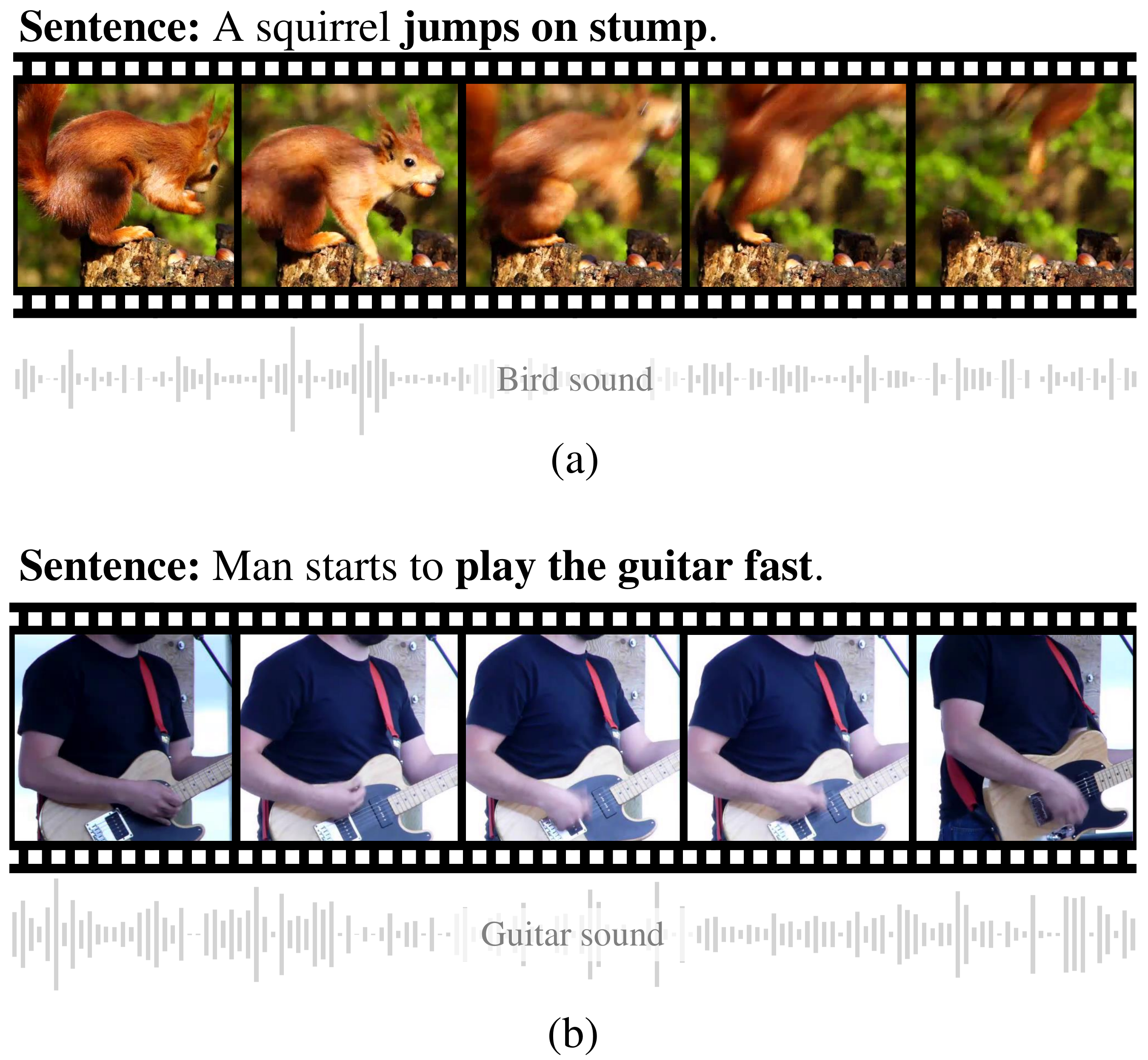}
\caption{Examples of video aided unsupervised grammar induction. We aim to improve the constituency parser by leveraging aligned video-sentence pairs.}
\label{fig:illustration}
\end{figure}

Previous methods~\cite{shi2019visually,kojima2020learned,zhao2020visually,jin2020grounded} learn to parse sentences by exploiting object information from images.
However, images are static and cannot present the dynamic interactions among visual objects, which usually correspond to verb phrases that carry important information.
% Therefore, relying on only image features may not be representative enough for matching with spans in certain circumstances.
Therefore, images and their descriptions may not be fully-representative of all linguistic phenomena encountered in learning, especially when action verbs are involved.
For example, as shown in Figure~\ref{fig:illustration}(a), when parsing a sentence ``A squirrel \textbf{jumps on stump}'', a single image cannot present the verb phrase ``\textbf{jumps on stump}'' accurately.
Moreover, as shown in Figure~\ref{fig:illustration}(b), 
%when parsing a sentence ``Man starts to \textbf{play the guitar fast}'',
the guitar sound and the moving fingers clearly indicate the speed of music playing, while it is impossible to present only with a static image as well.
Therefore, it is difficult for previous methods to learn these constituents, as static images they consider lack dynamic visual and audio information.

In this paper, we address this problem by leveraging video content to improve an unsupervised grammar induction model.
In particular, we exploit the current state-of-the-art techniques in both video and audio understanding, domains of which include object, motion, scene, face, optical character, sound, and speech recognition.
We extract features from their corresponding state-of-the-art models and analyze their usefulness with the VC-PCFG model~\cite{zhao2020visually}.
% baseline, the best performing model for image-grounded grammar induction so far.
Since different modalities may correlate with each other, independently modeling each of them may be sub-optimal.
We also propose a novel model, Multi-Modal Compound Probabilistic Context-Free Grammars (MMC-PCFG), to better model the correlation among these modalities.

Experiments on three benchmarks show substantial improvements when using each modality of the video content.
Moreover, our MMC-PCFG model that integrates information from different modalities further improves the overall performance.
Our code is available at~\url{https://github.com/Sy-Zhang/MMC-PCFG}.

The main contributions of this paper are:

\begin{itemize}
    \item We are the first to address video aided unsupervised grammar induction and demonstrate that verb related  features extracted from videos are beneficial to parsing.
    \item We perform a thorough analysis on different modalities of video content and propose a model to  effectively integrate these important modalities to train better constituency parsers.
    \item Experiments results demonstrate the effectiveness of our model over the previous state-of-the-art methods.
    %Our code and models are publicly available at~\url{https://github.com/XXX/XXX}.
\end{itemize}

\section{Background and Motivation}

Our model is motivated by C-PCFG~\cite{kim2019compound} and its variant of the image-aided unsupervised grammar induction model, VC-PCFG~\cite{zhao2020visually}. We will first review the evolution of these two frameworks in Sections~\ref{sec:c-pcfg}--\ref{sec:vc-pcfg}, and then discuss their limitations in Section~\ref{sec:limitation}.

\subsection{Compound PCFGs}
\label{sec:c-pcfg}

A probabilistic context-free grammar (PCFG) in Chomsky normal form can be defined as a 6-tuple $(S, \mathcal{N}, \mathcal{P}, \Sigma, \mathcal{R}, \Pi)$, where $S$ is the start symbol, $\mathcal{N}, \mathcal{P} \text{ and } \Sigma$ are the set of nonterminals, preterminals and terminals, respectively. $\mathcal{R}$ is a set of production rules with their probabilities stored in $\Pi$, where the rules include binary nonterminal expansions and unary terminal expansions. Given a certain number of nonterminal and preterminal categories, a PCFG induction model tries to estimate rule probabilities. By imposing a sentence-specific prior on the distribution of possible PCFGs, the compound PCFG model \cite{kim2019compound} uses a mixture of PCFGs to model individual sentences in contrast to previous models \cite{jin2018unsupervised} where a corpus-level prior is used. Specifically in the generative story, the rule probability $\pi_r$ is estimated by the model $g$ with a latent representation $\mathbf{z}$ for each sentence $\sigma$, which is in turn drawn from a prior $p(\mathbf{z})$:
%
% \begin{equation}
% \begin{aligned} 
% &S\rightarrow A,\qquad &A\in \mathcal{N},\\
% &A\rightarrow BC,\qquad &A\in \mathcal{N}, B, C\in \mathcal{N}\cup \mathcal{P},\\
% &T\rightarrow \omega,\qquad &T\in \mathcal{P}, \omega\in \Sigma.
% \end{aligned}
% \end{equation}
%
\begin{align}
    \pi_r=g_r(\mathbf{z};\theta),\qquad \mathbf{z}\sim p(\mathbf{z}).
\end{align}
The probabilities for the CFG initial expansion rules $S \rightarrow A$, nonterminal expansion rules $A \rightarrow B\ C$ and preterminal expansion rules $T \rightarrow w$ can be estimated by calculating scores of each combination of a parent category in the left hand side of a rule and all possible child categories in the right hand side of a rule:
\begin{equation}
\begin{aligned} 
    \pi_{S\rightarrow A}&=\frac{\exp(\mathbf{u}_A^\top f_s([\mathbf{w}_S;\mathbf{z}]))}{\sum_{A'\in \mathcal{N}} \exp(\mathbf{u}_{A'}f_s([\mathbf{w}_S;\mathbf{z}]))}, \\
    \pi_{A\rightarrow BC}&=\frac{\exp(\mathbf{u}_{BC}^\top [\mathbf{w}_A;\mathbf{z}])}{\sum_{B',C'\in \mathcal{N}\cup \mathcal{P}} \exp(\mathbf{u}_{B'C'}^\top [\mathbf{w}_A;\mathbf{z}]))},\\
    \pi_{T\rightarrow w}&=\frac{\exp(\mathbf{u}_w^\top f_t([\mathbf{w}_T;\mathbf{z}]))}{\sum_{w'\in \Sigma} \exp(\mathbf{u}_{w'}^Tf_t([\mathbf{w}_T;\mathbf{z}]))},
\end{aligned}
\end{equation}
where $A, B, C \in \mathcal{N}$ , $T \in \mathcal{P}$, $w \in \Sigma$, $\mathbf{w} \text{ and } \mathbf{u}$ vectorial representations of words and categories, and $f_t$ and $f_s$ are encoding functions such as neural networks. 

Optimization of the PCFG induction model usually involves maximizing the marginal likelihood of a training sentence $p(\sigma)$ for all sentences in a corpus. In the case of compound PCFGs:
\begin{equation}
    \log p_\theta(\sigma) = \log \int_{\mathbf{z}} \sum_{t \in \mathcal{T}_\mathcal{G}(\sigma)} p_\theta (t|\mathbf{z}) p(\mathbf{z}) d \mathbf{z},
\end{equation}
where $t$ is a possible binary branching parse tree of $\sigma$ among all possible trees $\mathcal{T}$ under a grammar $\mathcal{G}$. 
% Calculation of the marginal involves calculating the inner sum using the forward algorithm, and estimating the outer integral with Monte Carlo methods.
% Songyang: 这里我改了一点
Since computing the integral over $\mathbf{z}$ is intractable, $\log p_\theta(\sigma)$ can be optimized by maximizing its evidence lower bound $\mathrm{ELBO}(\sigma;\phi,\theta)$:
\begin{equation}
\begin{aligned}
    % &\log p_\theta(\sigma) \geq =\\
    \mathrm{ELBO}(\sigma;\phi,\theta)&=\mathbb{E}_{q_\phi(\mathbf{z}|\sigma)}[\log p_\theta(\sigma|\mathbf{z})]\\
    &-\mathrm{KL}[q_\phi (\mathbf{z}|\sigma)||p(\mathbf{z})],
\end{aligned}    
\end{equation}
where $q_\phi(\mathbf{z}|\sigma)$ is a variational posterior, a neural network parameterized with $\phi$. The sample log likelihood can be computed with the inside algorithm, while the KL term can be computed analytically when both prior $p(\mathbf{z})$  and the posterior approximation $q_\phi (\mathbf{z}|\sigma)$ are Gaussian~\cite{kingma2013auto}.

\subsection{Visualy Grounded Compound PCFGs}
\label{sec:vc-pcfg}

The visually grounded compound PCFGs (VC-PCFG) extends the compound PCFG model (C-PCFG) by including a matching model between images and text. The goal of the vision model is to match the representation of an image $\mathbf{v}$ to the representation of a span $\mathbf{c}$ in a parse tree $t$ of a sentence $\sigma$. 
% The word representation $\mathbf{h}_i$ for the $i$th word is calculated by a BiLSTM network, and the representation $\mathbf{c}$ of a particular span $c = w_i,\dots,w_j (0 < i < j \leq n)]$ is computed by applying a label-specific affine transformation $f_k$ to the averaged word representations in that span first, and summing over all the labels weighted by the probabilities of the phrasal labels given the span:
%
The word representation $\mathbf{h}_i$ for the $i$th word is calculated by a BiLSTM network.
Given a particular span $c = w_i,\dots,w_j (0 < i < j \leq n)]$, we then compute its representation $\mathbf{c}$.
We first compute the probabilities of its phrasal labels $\{p(k|c, \sigma)|1\leq k \leq K, K = |\mathcal{N}|\}$, as  described in Section \ref{sec:c-pcfg}. 
The representation $\mathbf{c}$ is the sum of all label-specific span representations weighted by the probabilities we predicted:

\begin{align}
    \mathbf{c} % = \sum_{k=1}^{K} p(k|c, \sigma)\mathbf{c}_k \\ 
    &=  \sum_{k=1}^{K} p(k|c, \sigma) f_k (\frac{1}{j-i+1} \sum_{l=i}^{j} \mathbf{h}_l),
\end{align}
%
% where the probabilities come from the parsing model described in Section \ref{sec:c-pcfg} and $K = |\mathcal{N}|$. 
Finally, the matching loss between a sentence $\sigma$ and an image representation $\mathbf{v}$ can be calculated as a sum over all matching losses between a span and the image representation, weighted by the marginal of a span from the parser:
\begin{equation}
    s_{img}(\mathbf{v}, \sigma) = \sum_{c\in\sigma} p(c|\sigma)h_{img}(\mathbf{c}, \mathbf{v}),
\label{eq:image_matching_loss}
\end{equation}
where $h_{img}(\mathbf{c}, \mathbf{v})$ is a hinge loss between the distances from the image representation $\mathbf{v}$ to the matching and unmatching (\textit{i.e.} sampled from a different sentence) spans $\mathbf{c}$ and $\mathbf{c}^\prime$, and the distances from the span $\mathbf{c}$ to the matching and unmatching (\textit{i.e.} sampled from a different image) image representations $\mathbf{v}$ and $\mathbf{v}^\prime$:
\begin{align}
    h_{img}(\mathbf{c},& \mathbf{v}) = \mathbb{E}_{\mathbf{c}^\prime}[\mathrm{cos}(\mathbf{c}^\prime, \mathbf{v}) - \mathrm{cos}(\mathbf{c}, \mathbf{v})) + \epsilon ]_+ \nonumber \\ 
    &+ \mathbb{E}_{\mathbf{v}^\prime}[\mathrm{cos}(\mathbf{c}, \mathbf{v}^\prime) - \mathrm{cos}(\mathbf{c}, \mathbf{v}) + \epsilon]_+,
\end{align}
where $\epsilon$ is a positive margin, and the expectations are approximated with one sample drawn from the training data.
During training, ELBO and the image-text matching loss are jointly optimized.

\noindent
\subsection{Limitation}
\label{sec:limitation}

VC-PCFG improves C-PCFG by leveraging the visual information from paired images. In their experiments~\cite{zhao2020visually}, comparing to C-PCFG, the largest improvement comes from NPs ($+11.9\%$ recall), while recall values of other frequent phrase types (VP, PP, SBAR, ADJP and ADVP) are fairly similar. The performance gain on NPs is also observed with another multi-modal induction model, VG-NSL~\cite{shi2019visually,kojima2020learned}.
Intuitively, image representations from image encoders trained on classification tasks very likely contain accurate information about objects in images, which is most relevant to identifying NPs%
\footnote{\citet{jin2020grounded} reports no improvement on English when incorporating visual information into a similar neural network-based PCFG induction model, which may be because \citet{zhao2020visually} removes punctuation from the training data, which removes a reliable source of phrasal boundary information. This loss is compensated by the induction model with image representations. We leave the study of evaluation configuration on induction results for future work.}%
.
However, they provide limited information for phrase types that mainly involve action and change, such as verb phrases.
Representations of dynamic scenes may help the induction model to identify verbs, and also contain information about the argument structure of the verbs and nouns based on features of actions and participants extracted from videos. 
% We also expect improvement on NPs as well, because videos may contain better exposure of certain objects described in the sentence.
%
Therefore, we propose a model that induces PCFGs from raw text aided by the multi-modal information extracted from videos, and expect to see accuracy gains on such places in comparison to the baseline systems.

\section{Multi-Modal Compound PCFGs}

In this section, we introduce the proposed multi-modal compound PCFGs (MMC-PCFG). Instead of purely relying on object information from images, we generalize VC-PCFG into the video domain, where multi-modal video information is considered. We first introduce the video representation in Section~\ref{sec:video-representation}. We then describe the procedure for matching the multi-modal video representation with each span in Section~\ref{sec:multi-modal-fusion}. After that we introduce the training and inference details in Section~\ref{sec:training-inference}.

\subsection{Video Representation}
\label{sec:video-representation}

A video contains a sequence of frames, denoted as ${V}=\{{v}_{i}\}_{i=1}^{L^0}$, where $v_{i}$ represents a frame in a video and $L^0$ indicates the total number of frames. We extract video representation from $M$ models trained on different tasks, which are called \emph{experts}. Each expert focuses on extracting a sequence of features of one type. In order to project different expert features into the same dimension, their feature sequences are feed into  linear layers (one per expert) with same output dimension. We denote the outputs of the $m$th expert after projection as $\mathbf{F}^m=\{\mathbf{f}^m_i\}_{i=1}^{L^m}$, where $\mathbf{f}^m_i$ and $L^m$ represent the $i$th feature and the total number of features of the $m$th expert, respectively.

A simple method would average each feature along the temporal dimension and then concatenating them together.
However, this would ignore the relations among different modalities and the temporal ordering within each modality. 
In this paper, we use a multi-modal transformer to collect video representations~\cite{gabeur2020multi,lei2020mart}.

The multi-modal transformer expects a sequence as input,
% , including input features $\mathbf{X}$, expert type embeddings $\mathbf{E}$ and positional encodings $\mathbf{P}$.
hence we concatenate all feature sequences together and take the form:
% The sequence of input features take the form,
\begin{equation}
\mathbf{X}=[\mathbf{f}_{avg}^1, \mathbf{f}_1^1, ..., \mathbf{f}_{L_1}^1,... \mathbf{f}_{avg}^M, \mathbf{f}_1^M, ...,  \mathbf{f}_{L_M}^M],    
\end{equation}
where $\mathbf{f}_{avg}^m$ is the averaged feature of $\{\mathbf{f}_i^m\}_{i=1}^{L_m}$. 
Each transformer layer has a standard architecture and consists of multi-head self-attention module and a feed forward network (FFN).
Since this architecture is permutation-invariant, we supplement it with expert type embeddings $\mathbf{E}$ and positional encoding $\mathbf{P}$ that are added to the input of each attention layer.
The expert type embeddings indicate the expert type for input features and take the form:
\begin{equation}
\mathbf{E}=[\mathbf{e}^1,\mathbf{e}^1,...,\mathbf{e}^1,...,\mathbf{e}^M,\mathbf{e}^M,...,\mathbf{e}^M],    
\end{equation}
where $\mathbf{e}^m$ is a learned embedding for the $m$th expert.
The positional encodings indicate the location of each feature within the video and take the form:
\begin{equation}
\mathbf{P}=[\mathbf{p}_0,\mathbf{p}_1,...,\mathbf{p}_{L_1},...,\mathbf{p}_0,\mathbf{p}_1,...,\mathbf{p}_{L_M}],    
\end{equation}
where fixed encodings are used~\cite{vaswani2017attention}.
After that, we collect the output of transformer that corresponds to the averaged features as the final video representation, \textit{i.e.}, $\bm{\Psi}=\{\bm{\psi}_{avg}^i\}_{i=1}^M$.
In this way, we can learn more effective video representation by modeling the correlations of features from different modalities and different timestamps.

\subsection{Video-Text Matching}
\label{sec:multi-modal-fusion}
To compute the similarity between a video $V$ and a particular span $c$, a span representation $\mathbf{c}$ is obtained following Section~\ref{sec:vc-pcfg} and projected to $M$ separate expert embeddings via gated embedding modules (one per expert)~\cite{miech2018learning}:%, 

\begin{equation}
    \begin{aligned}
        \bm{\xi}_1^i&=\bm{W}_1^i\bm{c}+\bm{b}_1^i, \\
        \bm{\xi}_2^i&=\bm{\xi}_1^i\circ sigmoid(\bm{W}_2^i\bm{\xi}_1^i+\bm{b}_2^i), \\
        \bm{\xi}^i&=\frac{\bm{\xi}_2^i}{\lVert \bm{\xi}_2^i \rVert_2},
    \end{aligned}
\end{equation}
where $i$ is the index of expert, $\bm{W}_1^i$, $\bm{W}_2^i$, $\bm{b}_1^i$, $\bm{b}_2^i$ are learnable parameters, $sigmoid$ is an element-wise sigmoid activation and $\circ$ is the element-wise multiplication.
We denote the set of expert embeddings as $\bm{\Xi}=\{\bm{\xi}^i\}_{i=1}^M$. The video-span similarity is computed as following,
\begin{equation}
\begin{aligned}
    \omega_i(\mathbf{c})&=\frac{\exp(\mathbf{u}_i^\top\mathbf{c})}{\sum_{j=1}^M\exp(\mathbf{u}_j^\top\mathbf{c})},\\
    o(\bm{\Xi},\bm{\Psi})&=\sum_{i=1}^M\omega_i(\mathbf{c})\mathrm{cos}(\bm{\xi}^i,\bm{\psi}^i),
\end{aligned}
\end{equation}
where $\{\mathbf{u}_i\}_{i=1}^M$ are learned weights. 
Given $\bm{\Xi}^\prime$, an unmatched span expert embeddings of $\bm{\Psi}$, and $\bm{\Psi}^\prime$, an unmatched video representation of $\bm{\Xi}$, the hinge loss for video is given by:
\begin{align}
    h_{vid}(\bm{\Xi},&\bm{\Psi}) = \mathbb{E}_{\mathbf{c}^\prime}[o(\bm{\Xi}^\prime, \bm{\Psi}) - o(\bm{\Xi}, \bm{\Psi})) + \epsilon ]_+ \nonumber \\ 
    &+ \mathbb{E}_{\bm{\Psi}^\prime}[o(\bm{\Xi}, \bm{\Psi}^\prime) - o(\bm{\Xi}, \bm{\Psi}) + \epsilon]_+,
\end{align}
where $\epsilon$ is a positive margin.
Finally the video-text matching loss is defined as:
\begin{equation}
    s_{vid}(V, \sigma) = \sum_{c\in\sigma} p(c|\sigma) h_{vid}(\bm{\Xi},\bm{\Psi}).
\end{equation}
Noted that $s_{vid}$ can be regarded as a generalized form of $s_{img}$ in Equation~\ref{eq:image_matching_loss}, where features from different timestamps and modalities are considered.

\subsection{Training and Inference}
\label{sec:training-inference}

During training, our model is optimized by the $\mathrm{ELBO}$ and the video-text matching loss:
\begin{equation}
    \mathcal{L}(\phi,\theta)=\sum_{(V,\sigma)\in \Omega}-\mathrm{ELBO}(\sigma;\phi,\theta)+\alpha s_{vid}(V,\sigma),
\end{equation}
where $\alpha$ is a hyper-parameter balancing these two loss terms and $\Omega$ is a video-sentence pair.

During inference, we predict the most likely tree $t^*$ given a sentence $\sigma$ without accessing videos. Since computing the integral over $\mathbf{z}$ is intractable, $t^*$ is estimated with the following approximation,
\begin{equation}
\begin{aligned}
    t^*&=\argmax_t\int_{\mathbf{z}} p_\theta (t|\mathbf{z}) p_\theta(\mathbf{z}|\sigma) d \mathbf{z}\\
    &\approx \argmax_t p_\theta (t|\sigma,\bm{\mu}_{\phi}(\sigma)),
\end{aligned}
\end{equation}
where $\bm{\mu}_{\phi}(\sigma)$ is the mean vector of the variational posterior $q_\phi(\mathbf{z}|\sigma)$ and $t^*$ can be obtained using the CYK algorithm~\cite{cocke1969programming,younger1967recognition,kasami1966efficient}.

% \section{Background and Motivation}

% In this paper, we extend the Visually Grounded Compound PCFGs(VC-PCFG)~\cite{zhao2020visually} to the video domain. We will introduce the relevant aspects of 
% \subsection{Compound PCFGs}

% \subsection{Visually grounded compound PCFGs}

% \subsection{Limitations of the VC-PCFG framework}

% \section{Multi-Modal compound PCFGs}

\section{Experiments}

\subsection{Datasets}

\noindent\textbf{DiDeMo~\cite{hendricks17iccv}}
% DiDeMo is originally designed for the task of moment localization with natural language.
 collects $10$K unedited, personal videos from Flickr with roughly $3-5$ pairs of descriptions and distinct moments per video.
% The videos are collected in an open-world setting and are diverse in content, such as pets, concerts, and sports games.
% We keep sentences with fewer than $20$ words in the training set due to the computational limitation. After filtering,
There are $\numprint{32994}$, $\numprint{4180}$ and $\numprint{4021}$ video-sentence pairs, validation and testing split.
% We split the dataset following the practice proposed by~\citet{albanie2020end}.

\noindent\textbf{YouCook2~\cite{ZhXuCoAAAI18}}
includes $2$K long untrimmed videos from $89$ cooking recipes. On average, each video has $6$ procedure steps described by imperative sentences.
% We keep  in our experiments due to the computational limitation. 
% After selecting sentences with fewer than $20$ words in the training set, 
There are $\numprint{8713}$, $\numprint{969}$ and $\numprint{3310}$ video-sentence pairs in the training, validation and testing sets. %where the training set covers $98.6\%$ samples of its original training set.

\noindent\textbf{MSRVTT~\cite{xu2016msr}} 
contains $10$K videos sourced from YouTube which are accompanied by $200$K descriptive
captions.
% We keep sentences with fewer than $20$ words in our experiments due to the computational limitation. 
% After selecting sentences with fewer than $20$ words, 
There are $\numprint{130260}$, $\numprint{9940}$ and $\numprint{59794}$ video-sentence pairs in the training, validation and testing sets.% which covers $97.0\%$ samples of the original dataset.

\subsection{Evaluation}

Following the evaluation practice in~\citet{zhao2020visually}, we discard punctuation and ignore trivial single-word and sentence-level spans at test time.
The gold parse trees are obtained by applying a state-of-the-art constituency parser, Benepar~\cite{Kitaev-2018-SelfAttentive}, on the testing set.
All models are run $4$ times for $10$ epochs with different random seeds. 
We evaluate both averaged corpus-level F1 (C-F1) and averaged sentence-level F1 (S-F1) numbers as well as their standard deviations.

\subsection{Expert Features}
In order to capture the rich content from videos, we extract features from the state-of-the-art models of different tasks, including object, action, scene, sound, face, speech, and optical character recognition (OCR). For object and action recognition, we explore multiple models with different architectures and pre-trained dataset. Details are as follows:

\noindent\textbf{Object features}
are extracted by two models: ResNeXt-$101$~\cite{xie2017aggregated}, pre-trained on Instagram hashtags~\cite{mahajan2018exploring} and fine-tuned on ImageNet~\cite{krizhevsky2012imagenet}, and SENet-$154$~\cite{hu2018senet}, trained on ImageNet. 
% For both models, video frames are extracted at $25$ fps and resized to $224\times 224$ pixels. 
These datasets include images of common objects, such as, ``cock'', ``kite'', and ``goose'', etc.
We use the predicted logits as object features for both models, where the dimension is $1000$.

\noindent\textbf{Action features}
are extracted by three models:
I3D trained on Kinetics-$400$~\cite{carreira2017quo}, R2P1D~\cite{tran2018closer} trained on IG-65M~\cite{ghadiyaram2019large} and S3DG~\cite{miech2020end} trained on HowTo100M~\cite{miech2019howto100m}. 
These datasets include videos of human actions, such as ``playing guitar'', ``ski jumping'', and ``jogging'', etc.
Following the same processing steps in their original work, we extract the predicted logits as action features, where the dimension is $400$ (I3D), $359$ (R2P1D) and $512$ (S3DG), respectively. 

\noindent\textbf{Scene features}
are extracted by DenseNet-$161$~\cite{huang2017densely} trained on Places$365$~\cite{zhou2017places}. 
Places$365$ contains images of different scenes, such as ``library'', ``valley'', and ``rainforest'', etc.
The predicted logits are used as scene features, where the feature dimension is $365$.

\noindent\textbf{Audio features}
are extracted by VGGish trained on YouTube-$8M$~\cite{hershey2017cnn}, where the feature dimension is $128$.
YouTube-$8M$ is a video dataset where different types of sound are involved, such as ``piano'', ``drum'', and ``violin''.

\noindent\textbf{OCR features}
are extracted by two steps: characters are first recognized by combining text detector Pixel Link~\cite{deng2018pixellink} and text recognizer SSFL~\cite{liu2018synthetically}. The characters are then converted to word embeddings through $word2vec$~\cite{mikolov2013efficient} as the final OCR features, where the feature dimension is $300$.

\noindent\textbf{Face features}
are extracted by combining face detector SSD ~\cite{liu2016ssd} and face recognizer ResNet50~\cite{he2016identity}. The feature dimension is $512$.

\noindent\textbf{Speech features}
are extracted by two steps: transcripts are first obtained via Google Cloud Speech to Text API. The transcripts are then converted to word embeddings through $word2vec$~\cite{mikolov2013efficient} as the final speech features, where the dimension is $300$.

\subsection{Implementation Details}
% \noindent\textbf{Dataset-specific Details.}
% \noindent\textbf{Training Details.} 

We keep sentences with fewer than $20$ words in the training set due to the computational limitation. After filtering, the training sets cover $99.4\%$, $98.5\%$ and $97.1\%$ samples of their original splits in DiDeMo, YouCook2 and MSRVTT.

We train baseline models, C-PCFG and VC-PCFG, with same hyper parameters suggested in~\citet{kim2019compound,zhao2020visually}.
Our MMC-PCFG is composed of a parsing model and a video-text matching model. The parsing model has the same parameters as  VC-PCFG (please refer to their paper for details).
% In more details, it has $60$ preterminals and $30$ nonterminals, represented by $256$-dimensional vectors. 
% The inference model first feeds $512$-dimensional word embeddings to a single-layer BiLSTM. After that, we apply a max-pooling layer over the hidden states and feed the output to a fully connected layer. Finally, we obtain $64$-dimensional mean and log-variances vectors.
% The span representation model is also a single-layer BiLSTM which has same hyper parameters with the inference model.
For video-text matching model, all extracted expert features are projected to $512$-dimensional vectors. The transformer has $2$ layers, a dropout probability of $10\%$, a hidden size of $512$ and an intermediate size of $2048$.
We select the top-$2000$ most common words as vocabulary for all datasets. 
All the baseline methods and our models are optimized using Adam~\cite{kingma2014adam} with the learning rate set to $0.001$, $\beta_1=0.75$ and $\beta_2=0.999$. All parameters are initialized with Xavier uniform initializer~\cite{glorot2010understanding}.
The batch size is set to $16$.

Due to the long video durations, it is infeasible to feed all features into the multi-modal transformer.
% An average pooling aggregation may lose the inherent temporal information.
Therefore, each feature from object, motion and scene categories is partitioned into $8$ chunks and then average-pooled within each chunk.
For features from other categories, global average pooling is applied.
In this way, the coarse-grained temporal information is preserved.
Noted that some videos do not have audio and some videos do not have detected faces or text characters. For these missing features, we pad them with zeros.
% In addition, since videos in YouCook2 are all recorded in kitchen and people or show faces, scene, face and speech features are not used in this dataset.
All the aforementioned expert features are obtained from~\citet{albanie2020end}.

\begin{table*}[t]
\setlength{\tabcolsep}{4.0pt}
\small
    \centering
    \begin{tabular}{ccccccccccc}
    \toprule
	\multicolumn{2}{c}{\multirow{2}*{Method}} & \multicolumn{5}{c}{DiDeMo} & \multicolumn{2}{c}{YouCook2}  & \multicolumn{2}{c}{MSRVTT} \\
	\cmidrule(lr){3-7}
	\cmidrule(lr){8-9}
	\cmidrule(lr){10-11}
	\multicolumn{2}{c}{} & NP & VP & PP & C-F1 & S-F1 & C-F1 & S-F1 & C-F1 & S-F1 \\
    \hline
    \multicolumn{2}{c}{LBranch}& $41.7$ & $0.1 $ & $0.1 $ & $16.2$ & $18.5$ & $6.8 $ & $5.9 $ & $14.4$ & $16.8$ \\
    \multicolumn{2}{c}{RBranch}& $32.8$ & $\mathbf{91.5}$ & $\mathbf{66.5}$ & $\mathit{53.6}$ & $\mathit{57.5}$ & $35.0$ & $41.6$ & $54.2$ & $58.6$ \\
    \multicolumn{2}{c}{Random} & $36.5_{\pm0.6 }$ & $30.5_{\pm0.5 }$ & $30.1_{\pm0.5 }$ & $29.4_{\pm0.3 }$ & $32.7_{\pm0.5 }$ & $21.2_{\pm0.2}$ & $24.0_{\pm0.2}$ & $27.2_{\pm0.1}$ & $30.5_{\pm0.1}$ \\
    \multicolumn{2}{c}{C-PCFG} & $\mathbf{72.9}_{\pm5.5 }$ & $16.5_{\pm6.2 }$ & $23.4_{\pm16.9}$ & $38.2_{\pm5.0 }$ & $40.4_{\pm4.1 }$ & $37.8_{\pm6.7}$ & $41.4_{\pm6.6}$ & $50.7_{\pm3.2}$ & $55.0_{\pm3.2}$ \\
    \hline
    \multirow{11}{*}{\rotatebox{90}{VC-PCFG}} & ResNeXt & $64.4_{\pm21.4}$ & $25.7_{\pm17.7}$ & $34.6_{\pm25.0}$ & $40.0_{\pm13.7}$ & $41.8_{\pm14.0}$ & $38.2_{\pm8.3}$ & $42.8_{\pm8.4}$ & $50.7_{\pm1.7}$ & $54.9_{\pm2.2}$ \\
    & SENet & $\mathit{70.5}_{\pm15.3}$ & $25.7_{\pm15.9}$ & $36.5_{\pm24.6}$ & $42.6_{\pm10.4}$ & $44.0_{\pm10.4}$ & $39.9_{\pm8.7}$ & $44.9_{\pm8.3}$ & $52.2_{\pm1.2}$ & $56.0_{\pm1.6}$ \\
    & I3D & $57.9_{\pm13.5}$ & $45.7_{\pm14.1}$ & $45.8_{\pm17.2}$ & $45.1_{\pm6.0 }$ & $49.2_{\pm6.0 }$ & $40.6_{\pm3.6}$ & $45.7_{\pm3.2}$ & $54.5_{\pm1.6}$ & $\mathit{59.1}_{\pm1.7}$ \\
    & R2P1D & $61.2_{\pm8.5 }$ & $38.1_{\pm5.4 }$ & $62.1_{\pm4.1 }$ & $48.1_{\pm4.4 }$ & $50.7_{\pm4.2 }$ & $39.4_{\pm8.1}$ & $44.4_{\pm8.3}$ & $54.0_{\pm2.5}$ & $58.0_{\pm2.3}$ \\
    & S3DG & $61.3_{\pm13.4}$ & $31.7_{\pm16.7}$ & $51.8_{\pm8.0 }$ & $44.0_{\pm2.7 }$ & $46.5_{\pm5.1 }$ & $39.3_{\pm6.5}$ & $44.1_{\pm6.6}$ & $50.7_{\pm3.2}$ & $54.7_{\pm2.9}$ \\
    & Scene & $62.2_{\pm9.6 }$ & $30.6_{\pm12.3}$ & $41.1_{\pm24.8}$ & $41.7_{\pm6.5 }$ & $44.9_{\pm7.4 }$ & $-$ & $-$ & $\mathit{54.6}_{\pm1.5}$ & $58.4_{\pm1.3}$ \\
    & Audio & $64.2_{\pm18.6}$ & $21.3_{\pm26.5}$ & $34.7_{\pm11.0}$ & $38.7_{\pm3.7 }$ & $39.5_{\pm5.2 }$ & $39.2_{\pm4.7}$ & $43.3_{\pm4.9}$ & $52.8_{\pm1.3}$ & $56.7_{\pm1.4}$ \\
    & OCR & $64.4_{\pm15.0}$ & $27.4_{\pm19.5}$ & $42.8_{\pm31.2}$ & $41.9_{\pm16.9}$ & $44.6_{\pm17.5}$ & $38.6_{\pm5.5}$ & $43.2_{\pm5.6}$ & $51.0_{\pm3.0}$ & $55.5_{\pm3.0}$ \\
    & Face & $60.8_{\pm16.0}$ & $31.5_{\pm17.0}$ & $52.8_{\pm9.8 }$ & $43.9_{\pm4.5 }$ & $46.3_{\pm5.5 }$ & $-$ & $-$ & $50.5_{\pm2.6}$ & $54.5_{\pm2.6}$ \\
    & Speech & $61.8_{\pm12.8}$ & $26.6_{\pm17.6}$ & $43.8_{\pm34.5}$ & $40.9_{\pm16.0}$ & $43.1_{\pm16.1}$ & $-$ & $-$ & $51.7_{\pm2.6}$ & $56.2_{\pm2.5}$ \\
    & Concat & $68.6_{\pm8.6 }$ & $24.9_{\pm19.9}$ & $39.7_{\pm19.5}$ & $42.2_{\pm12.3}$ & $43.2_{\pm14.2}$ & $\mathit{42.3}_{\pm5.7}$ & $\mathit{47.0}_{\pm5.6}$ & $49.8_{\pm4.1}$ & $54.2_{\pm4.0}$ \\
    \midrule
    \multicolumn{2}{c}{\textbf{MMC-PCFG}} & $67.9_{\pm9.8 }$ & $\mathit{52.3}_{\pm9.0 }$ & $\mathit{63.5}_{\pm8.6 }$ & $\mathbf{55.0}_{\pm3.7 }$ & $\mathbf{58.9}_{\pm3.4 }$ & $\mathbf{44.7}_{\pm5.2}$ & $\mathbf{48.9}_{\pm5.7}$ & $\mathbf{56.0}_{\pm1.4}$ & $\mathbf{60.0}_{\pm1.2}$ \\
    \bottomrule
    \end{tabular}
    \caption{Performance comparison on three benchmark datasets.}
    \label{tab:performance}
\end{table*}

\begin{figure*}[t]
\centering
\includegraphics[width=\textwidth]{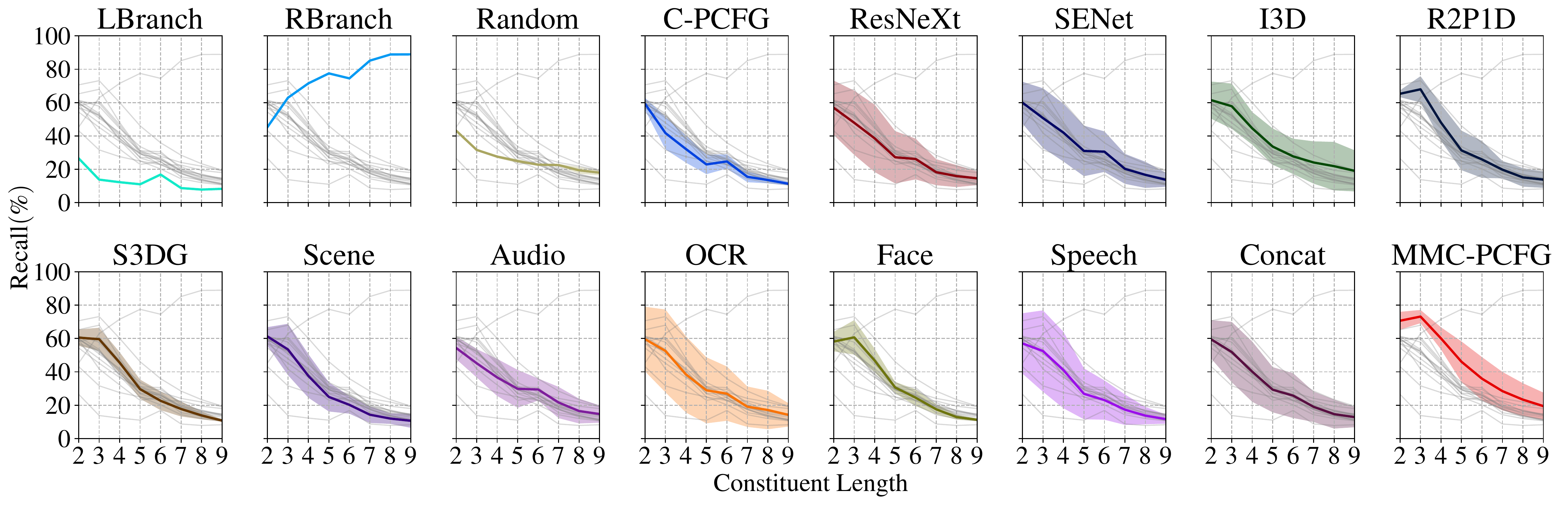}
\caption{Recall comparison over constituent length on DiDeMo. Methods are differentiated with colors. For easy comparison, we additional draw \textcolor{gray}{gray} lines in each figure to indicate the average recall shown by other figures.}
\label{fig:recall_vs_length}
\end{figure*}

\subsection{Main Results}

We evaluate the proposed MMC-PCFG approach on three datasets, and compare it with recently proposed state-of-the-art methods, C-PCFG~\cite{kim2019compound} and VC-PCFG~\cite{zhao2020visually}.
The results are summarized in Table~\ref{tab:performance}.
%~\ref{tab:DiDeMo}--~\ref{tab:YouCook2}.
The values high-lighted by \textbf{bold} and \textit{italic} fonts indicate the top-$2$ methods, respectively. All results are reported in percentage ($\%$).
LBranch, RBranch and Random represent left branching trees, right branching trees and random trees, respectively.
Since VC-PCFG is originally designed for images, it is not directly comparable with our method.
In order to allow VC-PCFG to accept videos as input, we average video features in the temporal dimension first and then feed them into the model.
We evaluate VC-PCFG with $10$, $7$, and $10$ expert features for DiDeMo, YouCook2 and MSRVTT, respectively.
In addition, we also include the concatenated averaged features (Concat).
Since object and action categories involve more than one expert, we directly use experts' names instead of their categories in Table~\ref{tab:performance}.

\noindent
\textbf{Overall performance comparison.}
We first compare the overall performance, \textit{i.e.}, C-F1 and S-F1, among all models, as shown in Table~\ref{tab:performance}.
The right branching model serves as a strong baseline, since English is a largely right-branching language.
C-PCFG learns parsing purely based on text.
Compared to C-PCFG, the better overall performance of VC-PCFG demonstrates the effectiveness of leveraging video information.
Compared within VC-PCFG, concatenating all features together may not even outperform a model trained on a single expert (R2P1D \textit{v.s.} Concat in DiDeMo and MSRVTT). The reason is that each expert is learned independently, where their correlations are not considered.
In contrast, our MMC-PCFG outperforms all baselines on C-F1 and S-F1 in all datasets.
The superior performance indicates that our model can leverage the benefits from all the experts\footnote{The larger improvement on DiDeMo may be caused by the diversity of the video content. Videos in DiDeMo are more diverse in scenes, actions and objects, which provide a great opportunity for leveraging video information.
%In contrast, YouCook2 is only comprised of imparative sentences and cooking videos, where objects and actions are limited.
}.
Moreover, the superior performance over Concat demonstrates the importance of modeling relations among different experts and different timestamps. 

\noindent
\textbf{Performance comparison among different phrase types.}
We compare the models' recalls on top-$3$ frequent phrase types (NP, VP and PP). These three types cover $77.4\%$, $80.1\%$ and $82.4\%$ spans of gold trees on DiDeMo, YouCook2 and MSRVTT, respectively. In the following, we compare their performance on DiDeMo, as shown in Table~\ref{tab:performance}.
Comparing VC-PCFG trained with a single expert, we find that object features (ResNeXt and SENet) achieve top-$2$ recalls on NPs, while action features (I3D, R2P1D and S3DG) achieve the top-$3$ recalls on VPs and PPs. 
It indicates that different experts help parser learn syntactic structures from different aspects.
Meanwhile, action features improve C-PCFG%
% \footnote{Experimentally, we find that C-PCFG tends to be left branching , which cause the lower recalls in VPs and PPs.} 
\footnote{The low performance of C-PCFG on DiDeMo in terms of VP recall may be caused by it attaching a high attaching PP to the rest of the sentence instead of the rest of the verb phrase, which breaks the whole VP. For PPs, C-PCFG attaches prepositions to the word in front, which may be caused by confusion between prepositions in PPs and phrasal verbs.}
on VPs and PPs by a large margin, which once again verifies the benefits of using video information.
% In the following, we mainly discuss on DiDeMo for two reasons:
% First, Videos in DiDeMo involve , while Videos in YouCook2 are limited cooking recorded kitchen.
% C-PCFG achieves high recall on NP, while perform much worse on VP and PP.

% Comparing VC-PCFG trained with different experts, 
% For NP, C-PCFG achieves the best performance in 
% Scene and SENet achieve the best recall in DiDeMo and YouCook2 respectively.
% For VP, SENet and Scene achieves top-$2$ recalls, while in YouCook2, the top-$2$ recalls are achieved by Audio and I3D.
% This is not surprising, since videos in DiDeMo are various in scenes (snowfield, beach, freeway \textit{etc.}), while videos in YouCook2 involves fine-grained objects (pan, plate, bread, \textit{etc.}) and actions (grilling, washing, cutting, \textit{etc.}) in kitchen.
% The reason is that YouCook2 involves many fine-grained actions , which is largely covered by YouTube-8M for sound recognition and .
% Comparing SENet154 and ResNeXt on NP, the better performance of SENet154 is also consistent with their performance on object recognition reported by~\citet{hu2018senet}.
% We also find that S3DG achieves comparable result in YouCook2, while it performs much worse in DiDeMo. 
% This is due to its pre-trained dataset, HowTo100M. This dataset are all instructional videos, including cooking, hand crafting, personal care, gardening or fitness, which is not various in scenes.
% In contrast,
Comparing our MMC-PCFG with VC-PCFG, our model achieves the top-$2$ recall and is smaller in variance in NP, VP and PP.
It demonstrates that our model can take the advantages of different experts and learn consistent grammar induction.
% From the experiment, we can observe that the 

\begin{figure}[t!]
\centering
\includegraphics[width=0.49\textwidth]{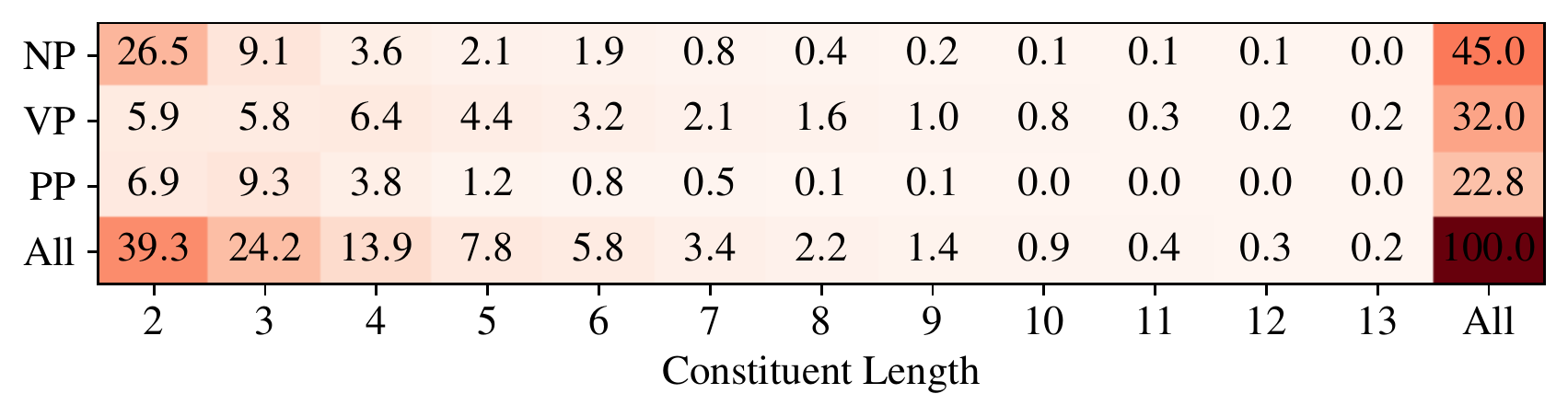}
\caption{Label distributions over the constituent length on DiDeMo. \textit{All} represent frequencies of constituent lengths.}
\label{fig:label_distribution}
\end{figure}

\subsection{Ablation Study}
\begin{figure}[t]
\centering
\includegraphics[width=0.48\textwidth]{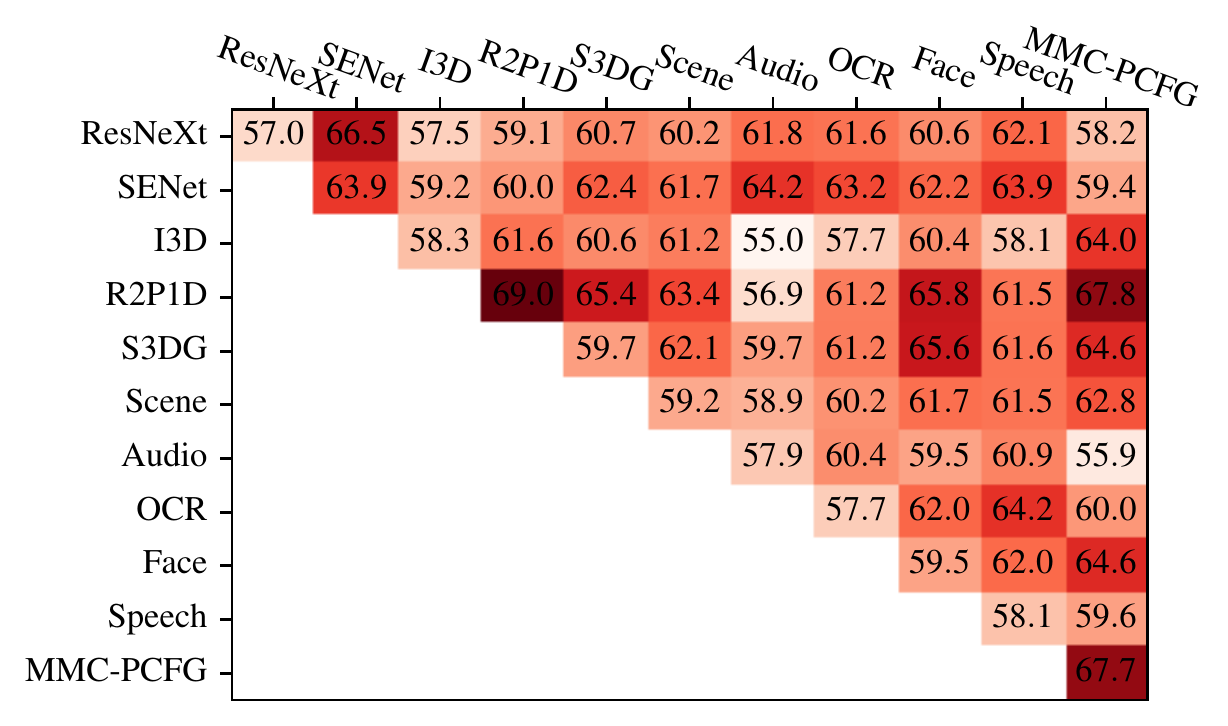}
\caption{Consistency scores for different models on DiDeMo. }
\label{fig:consistency_matrix}
\end{figure}

In this section, we conduct several ablation studies on DiDeMo, shown in Figures~\ref{fig:recall_vs_length}--\ref{fig:consistency_matrix}. All results are reported in percentage ($\%$).

\noindent
\textbf{Performance comparison over constituent length.}
We first demonstrate the model performance for constituents at different lengths in Figure~\ref{fig:recall_vs_length}.
As constituent length becomes longer, the recall of all models (except RBranch) decreases as expected~\cite{kim2019compound,zhao2020visually}.
MMC-PCFG outperforms C-PCFG and VC-PCFG under all constituent lengths.
We further illustrate the label distribution over constituent length in Figure~\ref{fig:label_distribution}.
We find that approximately $98.1\%$ of the constituents have fewer than $9$ words and most of them are NPs, VPs and PPs.
This suggests that the improvement on NPs, VPs and PPs can strongly affect the overall performance.

\noindent
\textbf{Consistency between different models.}
Next, we analyze the consistency of these different models. The consistency between two models is measured by averaging sentence-level F1 scores over all possible pairings of different runs\footnote{Different runs represent models trained with different seeds.}~\cite{williams2018latent}.
We plot the consistency for each pair of models in Figure~\ref{fig:consistency_matrix} and call it consistency matrix.
Comparing the self F1 of all the models (the diagonal in the matrix), R2P1D has the highest score, suggesting that R2P1D is the most reliable feature that can help parser to converge to a specific grammar.
Comparing the models trained with different single experts, ResNeXt \textit{v.s.} SENet reaches the highest non-self F1, since they are both object features trained on ImageNet and have similar effects to the parser. 
We also find that the lowest non-self F1 comes from Audio \textit{v.s.} I3D, since they are extracted from different modalities (video \textit{v.s.} sound).
% and trained on datasets annotated by largely different labels (VPs \textit{v.s.} NPs).
Compared with other models, our model is most consistent with R2P1D, indicating that R2P1D contributes most to our final prediction.
% In contrast, learning from models trained on different datasets forces our MMC-PCFG generate a more comprehensive grammar.

\begin{table}
\setlength{\tabcolsep}{0.5pt}
\small
    \centering
    \begin{tabular}{cccccc}
    \toprule
     Model & NP & VP & PP & C-F1 & S-F1 \\
    \midrule
    full & $68.0_{\pm9.9}$ & $52.7_{\pm9.0}$ & $63.8_{\pm8.7}$ & $55.3_{\pm3.3}$ & $59.0_{\pm3.4}$ \\
    w/o audio & $69.3_{\pm8.0}$  & $41.8_{\pm11.0}$ & $45.3_{\pm20.2}$ & $48.7_{\pm6.2}$ & $52.0_{\pm6.5}$ \\
    w/o text & $68.5_{\pm13.7}$  & $38.8_{\pm16.9}$ & $57.0_{\pm20.4}$ & $49.6_{\pm10.4}$ & $52.0_{\pm11.1}$ \\
    w/o video & $64.3_{\pm4.4}$  & $28.1_{\pm7.5}$ & $38.9_{\pm25.6}$ & $41.4_{\pm6.0}$ & $44.8_{\pm5.9}$ \\
    \bottomrule
    \end{tabular}
    \caption{Performance comparison over modalities on MMC-PCFG on DiDeMo.}
    \label{tab:modalities}
\end{table}

\noindent
\textbf{Contribution of different modalities.}
We also evaluate how different modalities contribute to the performance of MMC-PCFG.
We divide current experts into three groups, video (objects, action, scene and face), audio (audio) and text (OCR and ASR). By ablating one group during training, we find that the model without video experts has the largest performance drops (see Table~\ref{tab:modalities}). Therefore, videos contribute most to the performance among all modalities.

\begin{figure}
    \centering
    \includegraphics[width=0.47\textwidth]{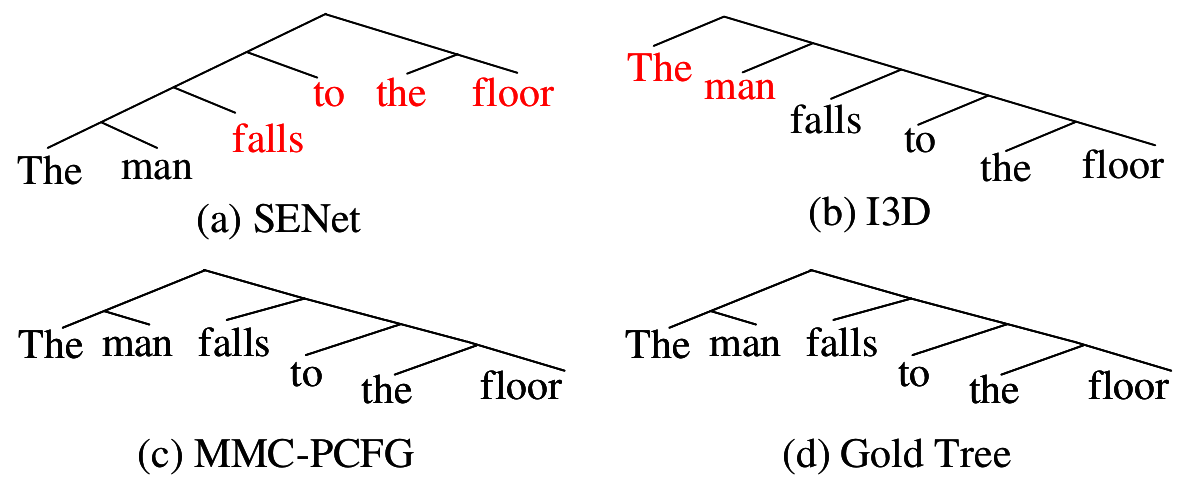}
    \caption{Parse trees predicted by different models for the sentence \textit{The man falls to the floor}.}
    \label{fig:example}
\end{figure}
\subsection{Qualitative Analysis}
In  Figure~\ref{fig:example}, we visualize a parse tree predicted by the best run of SENet154, I3D and MMC-PCFG. We can observe that SENet identifies all NPs but fails at the VP. I3D correctly predicts the VP but fails at recognizing a NP, ``the man''. Our MMC-PCFG can take advantages of all experts and produce the  correct prediction.

\section{Related Work}
\noindent\textbf{Grammar Induction}
Grammar induction and unsupervised parsing has been a long-standing problem in computational linguistics \citep{carroll1992two}. Recent work utilized neural networks in predicting constituency structures with no supervision~\cite{shen2018neural,drozdov2019unsupervised,shen2018ordered,kim2019compound,Jin2019-us} and showed promising results.
In addition to learning purely from text, there is a growing interest to use image information to improve accuracy of induced constituency trees~\cite{shi2019visually,kojima2020learned,zhao2020visually,jin2020grounded}.
Different from previous work, our work improves the constituency parser by using videos containing richer information than images.

\noindent\textbf{Video-Text Matching}
Video-text matching has been widely studied in various tasks, such as video retrieval~\cite{Liu2019a,gabeur2020multi}, moment localization with natural language~\cite{zhang2019exploiting,2DTAN_2020_AAAI} and video question and answering~\cite{xu2017video,jin2019multi}.
It aims to learn video-semantic representation in a joint embedding space.
Recent works~\cite{Liu2019a,gabeur2020multi,chen2020learning} focus on learning video's multi-modal representation to match with text.
In this work, we borrow this idea to match video and textual representations.

\section{Conclusion}

In this work, we have presented a new task referred to as video-aided unsupervised grammar induction.
This task aims to improve grammar induction models by using aligned video-sentence pairs as an effective way to address the limitation of current image-based methods where only object information from static images is considered and important verb related information from vision is missing. 
Moreover, we present Multi-Modal Compound Probabilistic Context-Free Grammars (MMC-PCFG) to effectively integrate video features extracted from different modalities to induce more accurate grammars.
Experiments on three datasets demonstrate the effectiveness of our method.

\section{Acknowledgement}
We thank the support of NSF awards IIS-1704337, IIS-1722847, IIS-1813709, and the generous gift from our corporate sponsors.

\bibliography{reference}

\begin{thebibliography}{55}
\expandafter\ifx\csname natexlab\endcsname\relax\def\natexlab#1{#1}\fi

\bibitem[{Albanie et~al.(2020)Albanie, Liu, Nagrani, Miech, Coto, Laptev,
  Sukthankar, Ghanem, Zisserman, Gabeur et~al.}]{albanie2020end}
Samuel Albanie, Yang Liu, Arsha Nagrani, Antoine Miech, Ernesto Coto, Ivan
  Laptev, Rahul Sukthankar, Bernard Ghanem, Andrew Zisserman, Valentin Gabeur,
  et~al. 2020.
\newblock The end-of-end-to-end: A video understanding pentathlon challenge
  (2020).
\newblock \emph{arXiv preprint arXiv:2008.00744}.

\bibitem[{Carreira and Zisserman(2017)}]{carreira2017quo}
Joao Carreira and Andrew Zisserman. 2017.
\newblock Quo vadis, action recognition? a new model and the kinetics dataset.
\newblock In \emph{CVPR}.

\bibitem[{Carroll and Charniak(1992)}]{Carroll1992-jf}
Glenn Carroll and Eugene Charniak. 1992.
\newblock Two experiments on learning probabilistic dependency grammars from
  corpora.
\newblock \emph{Working Notes of the Workshop on Statistically-Based NLP
  Techniques}, (March):1--13.

\bibitem[{Chen et~al.(2020)Chen, Jiang, Liu, and Jiang}]{chen2020learning}
Shaoxiang Chen, Wenhao Jiang, Wei Liu, and Yu-Gang Jiang. 2020.
\newblock Learning modality interaction for temporal sentence localization and
  event captioning in videos.
\newblock In \emph{ECCV}.

\bibitem[{Cocke(1969)}]{cocke1969programming}
John Cocke. 1969.
\newblock \emph{Programming languages and their compilers: Preliminary notes}.
\newblock New York University.

\bibitem[{Deng et~al.(2018)Deng, Liu, Li, and Cai}]{deng2018pixellink}
Dan Deng, Haifeng Liu, Xuelong Li, and Deng Cai. 2018.
\newblock Pixellink: Detecting scene text via instance segmentation.
\newblock In \emph{AAAI}.

\bibitem[{Drozdov et~al.(2019)Drozdov, Verga, Yadav, Iyyer, and
  McCallum}]{drozdov2019unsupervised}
Andrew Drozdov, Patrick Verga, Mohit Yadav, Mohit Iyyer, and Andrew McCallum.
  2019.
\newblock Unsupervised latent tree induction with deep inside-outside recursive
  auto-encoders.
\newblock In \emph{NAACL}.

\bibitem[{Fried et~al.(2019)Fried, Kitaev, and Klein}]{fried-etal-2019-cross}
Daniel Fried, Nikita Kitaev, and Dan Klein. 2019.
\newblock Cross-domain generalization of neural constituency parsers.
\newblock In \emph{ACL}.

\bibitem[{Gabeur et~al.(2020)Gabeur, Sun, Alahari, and
  Schmid}]{gabeur2020multi}
Valentin Gabeur, Chen Sun, Karteek Alahari, and Cordelia Schmid. 2020.
\newblock Multi-modal transformer for video retrieval.
\newblock In \emph{ECCV}.

\bibitem[{Ghadiyaram et~al.(2019)Ghadiyaram, Tran, and
  Mahajan}]{ghadiyaram2019large}
Deepti Ghadiyaram, Du~Tran, and Dhruv Mahajan. 2019.
\newblock Large-scale weakly-supervised pre-training for video action
  recognition.
\newblock In \emph{CVPR}.

\bibitem[{Gleitman(1990)}]{Gleitman1990}
Lila Gleitman. 1990.
\newblock \href
  {http://www.ai.mit.edu/projects/dm/bp/gleitman90-verbmeaning.pdf} {{The
  Structural Sources of Verb Meanings}}.
\newblock \emph{Language Acquisition}, 1(1):3--55.

\bibitem[{Glorot and Bengio(2010)}]{glorot2010understanding}
Xavier Glorot and Yoshua Bengio. 2010.
\newblock Understanding the difficulty of training deep feedforward neural
  networks.
\newblock In \emph{AISTATS}.

\bibitem[{He et~al.(2016)He, Zhang, Ren, and Sun}]{he2016identity}
Kaiming He, Xiangyu Zhang, Shaoqing Ren, and Jian Sun. 2016.
\newblock Identity mappings in deep residual networks.
\newblock In \emph{ECCV}.

\bibitem[{Hendricks et~al.(2017)Hendricks, Wang, Shechtman, Sivic, Darrell, and
  Russell}]{hendricks17iccv}
Lisa~Anne Hendricks, Oliver Wang, Eli Shechtman, Josef Sivic, Trevor Darrell,
  and Bryan Russell. 2017.
\newblock Localizing moments in video with natural language.
\newblock In \emph{ICCV}.

\bibitem[{Hershey et~al.(2017)Hershey, Chaudhuri, Ellis, Gemmeke, Jansen,
  Moore, Plakal, Platt, Saurous, Seybold et~al.}]{hershey2017cnn}
Shawn Hershey, Sourish Chaudhuri, Daniel~PW Ellis, Jort~F Gemmeke, Aren Jansen,
  R~Channing Moore, Manoj Plakal, Devin Platt, Rif~A Saurous, Bryan Seybold,
  et~al. 2017.
\newblock {CNN} architectures for large-scale audio classification.
\newblock In \emph{ICASSP}.

\bibitem[{Hu et~al.(2018)Hu, Shen, and Sun}]{hu2018senet}
Jie Hu, Li~Shen, and Gang Sun. 2018.
\newblock Squeeze-and-excitation networks.
\newblock In \emph{CVPR}.

\bibitem[{Huang et~al.(2017)Huang, Liu, Van Der~Maaten, and
  Weinberger}]{huang2017densely}
Gao Huang, Zhuang Liu, Laurens Van Der~Maaten, and Kilian~Q Weinberger. 2017.
\newblock Densely connected convolutional networks.
\newblock In \emph{CVPR}.

\bibitem[{Jin et~al.(2018)Jin, Doshi-Velez, Miller, Schuler, and
  Schwartz}]{jin2018unsupervised}
Lifeng Jin, Finale Doshi-Velez, Timothy Miller, William Schuler, and Lane
  Schwartz. 2018.
\newblock Unsupervised grammar induction with depth-bounded pcfg.
\newblock \emph{TACL}.

\bibitem[{Jin et~al.(2019{\natexlab{a}})Jin, Doshi-Velez, Miller, Schwartz, and
  Schuler}]{Jin2019-us}
Lifeng Jin, Finale Doshi-Velez, Timothy Miller, Lane Schwartz, and William
  Schuler. 2019{\natexlab{a}}.
\newblock Unsupervised learning of {PCFGs} with normalizing flow.
\newblock In \emph{{ACL}}.

\bibitem[{Jin and Schuler(2020)}]{jin2020grounded}
Lifeng Jin and William Schuler. 2020.
\newblock Grounded pcfg induction with images.
\newblock In \emph{AACL-IJCNLP}.

\bibitem[{Jin et~al.(2019{\natexlab{b}})Jin, Zhao, Gu, Yu, Xiao, and
  Zhuang}]{jin2019multi}
Weike Jin, Zhou Zhao, Mao Gu, Jun Yu, Jun Xiao, and Yueting Zhuang.
  2019{\natexlab{b}}.
\newblock Multi-interaction network with object relation for video question
  answering.
\newblock In \emph{ACM Multimedia}.

\bibitem[{Kasami(1966)}]{kasami1966efficient}
Tadao Kasami. 1966.
\newblock An efficient recognition and syntax-analysis algorithm for
  context-free languages.
\newblock \emph{Coordinated Science Laboratory Report no. R-257}.

\bibitem[{Kim et~al.(2019)Kim, Dyer, and Rush}]{kim2019compound}
Yoon Kim, Chris Dyer, and Alexander~M Rush. 2019.
\newblock Compound probabilistic context-free grammars for grammar induction.
\newblock In \emph{ACL}.

\bibitem[{Kingma and Ba(2015)}]{kingma2014adam}
Diederik~P Kingma and Jimmy Ba. 2015.
\newblock Adam: A method for stochastic optimization.
\newblock In \emph{ICLR}.

\bibitem[{Kingma and Welling(2014)}]{kingma2013auto}
Diederik~P Kingma and Max Welling. 2014.
\newblock Auto-encoding variational bayes.
\newblock In \emph{ICLR}.

\bibitem[{Kitaev and Klein(2018)}]{Kitaev-2018-SelfAttentive}
Nikita Kitaev and Dan Klein. 2018.
\newblock Constituency parsing with a self-attentive encoder.
\newblock In \emph{ACL}.

\bibitem[{Kojima et~al.(2020)Kojima, Averbuch-Elor, Rush, and
  Artzi}]{kojima2020learned}
Noriyuki Kojima, Hadar Averbuch-Elor, Alexander~M Rush, and Yoav Artzi. 2020.
\newblock What is learned in visually grounded neural syntax acquisition.
\newblock In \emph{ACL}.

\bibitem[{Krizhevsky et~al.(2012)Krizhevsky, Sutskever, and
  Hinton}]{krizhevsky2012imagenet}
Alex Krizhevsky, Ilya Sutskever, and Geoffrey~E Hinton. 2012.
\newblock Imagenet classification with deep convolutional neural networks.
\newblock In \emph{NIPS}.

\bibitem[{Lei et~al.(2020)Lei, Wang, Shen, Yu, Berg, and Bansal}]{lei2020mart}
Jie Lei, Liwei Wang, Yelong Shen, Dong Yu, Tamara~L Berg, and Mohit Bansal.
  2020.
\newblock Mart: Memory-augmented recurrent transformer for coherent video
  paragraph captioning.
\newblock In \emph{ACL}.

\bibitem[{Liu et~al.(2016)Liu, Anguelov, Erhan, Szegedy, Reed, Fu, and
  Berg}]{liu2016ssd}
Wei Liu, Dragomir Anguelov, Dumitru Erhan, Christian Szegedy, Scott Reed,
  Cheng-Yang Fu, and Alexander~C Berg. 2016.
\newblock {SSD}: Single shot multibox detector.
\newblock In \emph{ECCV}.

\bibitem[{Liu et~al.(2019)Liu, Albanie, Nagrani, and Zisserman}]{Liu2019a}
Y.~Liu, S.~Albanie, A.~Nagrani, and A.~Zisserman. 2019.
\newblock Use what you have: Video retrieval using representations from
  collaborative experts.
\newblock In \emph{arXiv preprint arxiv:1907.13487}.

\bibitem[{Liu et~al.(2018)Liu, Wang, Jin, and Wassell}]{liu2018synthetically}
Yang Liu, Zhaowen Wang, Hailin Jin, and Ian Wassell. 2018.
\newblock Synthetically supervised feature learning for scene text recognition.
\newblock In \emph{ECCV}.

\bibitem[{Mahajan et~al.(2018)Mahajan, Girshick, Ramanathan, He, Paluri, Li,
  Bharambe, and van~der Maaten}]{mahajan2018exploring}
Dhruv Mahajan, Ross Girshick, Vignesh Ramanathan, Kaiming He, Manohar Paluri,
  Yixuan Li, Ashwin Bharambe, and Laurens van~der Maaten. 2018.
\newblock Exploring the limits of weakly supervised pretraining.
\newblock In \emph{ECCV}.

\bibitem[{Marcus et~al.(1993)Marcus, Santorini, and
  Marcinkiewicz}]{marcus-etal-1993-building}
Mitchell~P. Marcus, Beatrice Santorini, and Mary~Ann Marcinkiewicz. 1993.
\newblock Building a large annotated corpus of {E}nglish: The {P}enn
  {T}reebank.
\newblock \emph{Computational Linguistics}, 19(2):313--330.

\bibitem[{Miech et~al.(2020)Miech, Alayrac, Smaira, Laptev, Sivic, and
  Zisserman}]{miech2020end}
Antoine Miech, Jean-Baptiste Alayrac, Lucas Smaira, Ivan Laptev, Josef Sivic,
  and Andrew Zisserman. 2020.
\newblock End-to-end learning of visual representations from uncurated
  instructional videos.
\newblock In \emph{CVPR}.

\bibitem[{Miech et~al.(2018)Miech, Laptev, and Sivic}]{miech2018learning}
Antoine Miech, Ivan Laptev, and Josef Sivic. 2018.
\newblock Learning a text-video embedding from incomplete and heterogeneous
  data.
\newblock \emph{arXiv preprint arXiv:1804.02516}.

\bibitem[{Miech et~al.(2019)Miech, Zhukov, Alayrac, Tapaswi, Laptev, and
  Sivic}]{miech2019howto100m}
Antoine Miech, Dimitri Zhukov, Jean-Baptiste Alayrac, Makarand Tapaswi, Ivan
  Laptev, and Josef Sivic. 2019.
\newblock {HowTo100M}: Learning a text-video embedding by watching hundred
  million narrated video clips.
\newblock In \emph{ICCV}.

\bibitem[{Mikolov et~al.(2013)Mikolov, Chen, Corrado, and
  Dean}]{mikolov2013efficient}
Tomas Mikolov, Kai Chen, Greg Corrado, and Jeffrey Dean. 2013.
\newblock Efficient estimation of word representations in vector space.
\newblock \emph{arXiv preprint arXiv:1301.3781}.

\bibitem[{Pinker and MacWhinney(1987)}]{pinker1987bootstrapping}
Steven Pinker and B~MacWhinney. 1987.
\newblock {The bootstrapping problem in language acquisition}.
\newblock \emph{Mechanisms of language acquisition}, pages 399--441.

\bibitem[{Shen et~al.(2018{\natexlab{a}})Shen, Lin, wei Huang, and
  Courville}]{shen2018neural}
Yikang Shen, Zhouhan Lin, Chin wei Huang, and Aaron Courville.
  2018{\natexlab{a}}.
\newblock Neural language modeling by jointly learning syntax and lexicon.
\newblock In \emph{ICLR}.

\bibitem[{Shen et~al.(2018{\natexlab{b}})Shen, Tan, Sordoni, and
  Courville}]{shen2018ordered}
Yikang Shen, Shawn Tan, Alessandro Sordoni, and Aaron Courville.
  2018{\natexlab{b}}.
\newblock Ordered neurons: Integrating tree structures into recurrent neural
  networks.
\newblock In \emph{ICLR}.

\bibitem[{Shi et~al.(2019)Shi, Mao, Gimpel, and Livescu}]{shi2019visually}
Haoyue Shi, Jiayuan Mao, Kevin Gimpel, and Karen Livescu. 2019.
\newblock Visually grounded neural syntax acquisition.
\newblock In \emph{ACL}.

\bibitem[{Tomasello(2003)}]{Tomasello2003}
Michael Tomasello. 2003.
\newblock \emph{{Constructing a language: A usage-based theory of language
  acquisition.}}
\newblock Harvard University Press, Cambridge, MA, US.

\bibitem[{Tran et~al.(2018)Tran, Wang, Torresani, Ray, LeCun, and
  Paluri}]{tran2018closer}
Du~Tran, Heng Wang, Lorenzo Torresani, Jamie Ray, Yann LeCun, and Manohar
  Paluri. 2018.
\newblock A closer look at spatiotemporal convolutions for action recognition.
\newblock In \emph{CVPR}.

\bibitem[{Vaswani et~al.(2017)Vaswani, Shazeer, Parmar, Uszkoreit, Jones,
  Gomez, Kaiser, and Polosukhin}]{vaswani2017attention}
Ashish Vaswani, Noam Shazeer, Niki Parmar, Jakob Uszkoreit, Llion Jones,
  Aidan~N Gomez, {\L}ukasz Kaiser, and Illia Polosukhin. 2017.
\newblock Attention is all you need.
\newblock In \emph{NIPS}.

\bibitem[{Williams et~al.(2018)Williams, Drozdov*, and
  Bowman}]{williams2018latent}
Adina Williams, Andrew Drozdov*, and Samuel~R Bowman. 2018.
\newblock Do latent tree learning models identify meaningful structure in
  sentences?
\newblock \emph{TACL}.

\bibitem[{Xie et~al.(2017)Xie, Girshick, Doll{\'a}r, Tu, and
  He}]{xie2017aggregated}
Saining Xie, Ross Girshick, Piotr Doll{\'a}r, Zhuowen Tu, and Kaiming He. 2017.
\newblock Aggregated residual transformations for deep neural networks.
\newblock In \emph{CVPR}.

\bibitem[{Xu et~al.(2017)Xu, Zhao, Xiao, Wu, Zhang, He, and
  Zhuang}]{xu2017video}
Dejing Xu, Zhou Zhao, Jun Xiao, Fei Wu, Hanwang Zhang, Xiangnan He, and Yueting
  Zhuang. 2017.
\newblock Video question answering via gradually refined attention over
  appearance and motion.
\newblock In \emph{ACM Multimedia}.

\bibitem[{Xu et~al.(2016)Xu, Mei, Yao, and Rui}]{xu2016msr}
Jun Xu, Tao Mei, Ting Yao, and Yong Rui. 2016.
\newblock Msr-vtt: A large video description dataset for bridging video and
  language.
\newblock In \emph{CVPR}.

\bibitem[{Younger(1967)}]{younger1967recognition}
Daniel~H Younger. 1967.
\newblock Recognition and parsing of context-free languages in time n3.
\newblock \emph{Information and control}, 10(2):189--208.

\bibitem[{Zhang et~al.(2020)Zhang, Peng, Fu, and Luo}]{2DTAN_2020_AAAI}
Songyang Zhang, Houwen Peng, Jianlong Fu, and Jiebo Luo. 2020.
\newblock Learning 2d temporal adjacent networks formoment localization with
  natural language.
\newblock In \emph{AAAI}.

\bibitem[{Zhang et~al.(2019)Zhang, Su, and Luo}]{zhang2019exploiting}
Songyang Zhang, Jinsong Su, and Jiebo Luo. 2019.
\newblock Exploiting temporal relationships in video moment localization with
  natural language.
\newblock In \emph{ACM Multimedia}.

\bibitem[{Zhao and Titov(2020)}]{zhao2020visually}
Yanpeng Zhao and Ivan Titov. 2020.
\newblock Visually grounded compound {PCFG}s.
\newblock In \emph{EMNLP}.

\bibitem[{Zhou et~al.(2017)Zhou, Lapedriza, Khosla, Oliva, and
  Torralba}]{zhou2017places}
Bolei Zhou, Agata Lapedriza, Aditya Khosla, Aude Oliva, and Antonio Torralba.
  2017.
\newblock Places: A 10 million image database for scene recognition.
\newblock \emph{TPAMI}.

\bibitem[{Zhou et~al.(2018)Zhou, Xu, and Corso}]{ZhXuCoAAAI18}
Luowei Zhou, Chenliang Xu, and Jason~J Corso. 2018.
\newblock Towards automatic learning of procedures from web instructional
  videos.
\newblock In \emph{AAAI}.

\end{thebibliography}
\bibliographystyle{acl_natbib}

\newpage
\appendix

\onecolumn
\section{Performance Comparison - Full Tables}
\label{sec:appendix}

\begin{table*}[hbt!]
\small
    \centering
    \begin{tabular}{cccccccccc}
    \toprule
	\multicolumn{2}{c}{Method} & NP & VP & PP & SBAR & ADJP & ADVP & C-F1 & S-F1 \\
	\hline
    \multicolumn{2}{c}{LBranch} & $41.7$ & $0.1 $ & $0.1 $ & $0.7 $ & $7.2 $ & $0.0 $ & $16.2$ & $18.5$ \\
    \multicolumn{2}{c}{RBranch} & $32.8$ & $\mathbf{91.5}$ & $\mathbf{66.5}$ & $\mathbf{88.2}$ & $\mathbf{36.9}$ & $\mathbf{63.6}$ & $\mathit{53.6}$ & $\mathit{57.5}$ \\
    \multicolumn{2}{c}{Random}  & $36.5_{\pm0.6 }$ & $30.5_{\pm0.5 }$ & $30.1_{\pm0.5 }$ & $25.7_{\pm2.8 }$ & $29.5_{\pm2.3 }$ & $28.5_{\pm4.8 }$ & $29.4_{\pm0.3 }$ & $32.7_{\pm0.5 }$ \\
    \multicolumn{2}{c}{C-PCFG} & $\mathbf{72.9}_{\pm5.5 }$ & $16.5_{\pm6.2 }$ & $23.4_{\pm16.9}$ & $26.6_{\pm15.9}$ & $25.0_{\pm11.6}$ & $14.7_{\pm12.8}$ & $38.2_{\pm5.0 }$ & $40.4_{\pm4.1 }$ \\
    \hline
    \multirow{11}{*}{\rotatebox{90}{VC-PCFG}} & ResNeXt & $64.4_{\pm21.4}$ & $25.7_{\pm17.7}$ & $34.6_{\pm25.0}$ & $40.5_{\pm26.3}$ & $16.7_{\pm9.5 }$ & $28.4_{\pm21.3}$ & $40.0_{\pm13.7}$ & $41.8_{\pm14.0}$ \\
    & SENet & $70.5_{\pm15.3}$ & $25.7_{\pm15.9}$ & $36.5_{\pm24.6}$ & $36.8_{\pm25.9}$ & $21.2_{\pm12.5}$ & $23.6_{\pm16.8}$ & $42.6_{\pm10.4}$ & $44.0_{\pm10.4}$ \\
    & I3D & $57.9_{\pm13.5}$ & $45.7_{\pm14.1}$ & $45.8_{\pm17.2}$ & $38.2_{\pm14.8}$ & $28.4_{\pm9.2 }$ & $22.0_{\pm9.3 }$ & $45.1_{\pm6.0 }$ & $49.2_{\pm6.0 }$ \\
    & R2P1D & $61.2_{\pm8.5 }$ & $38.1_{\pm5.4 }$ & $62.1_{\pm4.1 }$ & $61.5_{\pm5.1 }$ & $21.4_{\pm11.4}$ & $40.8_{\pm7.3 }$ & $48.1_{\pm4.4 }$ & $50.7_{\pm4.2 }$ \\
    & S3DG & $61.3_{\pm13.4}$ & $31.7_{\pm16.7}$ & $51.8_{\pm8.0 }$ & $50.3_{\pm6.5 }$ & $18.0_{\pm4.5 }$ & $35.2_{\pm11.4}$ & $44.0_{\pm2.7 }$ & $46.5_{\pm5.1 }$ \\
    & Scene & $62.2_{\pm9.6 }$ & $30.6_{\pm12.3}$ & $41.1_{\pm24.8}$ & $35.2_{\pm21.9}$ & $21.4_{\pm14.0}$ & $27.6_{\pm17.1}$ & $41.7_{\pm6.5 }$ & $44.9_{\pm7.4 }$ \\
    & Audio & $64.2_{\pm18.6}$ & $21.3_{\pm26.5}$ & $34.7_{\pm11.0}$ & $37.3_{\pm19.6}$ & $26.1_{\pm4.9 }$ & $18.2_{\pm11.6}$ & $38.7_{\pm3.7 }$ & $39.5_{\pm5.2 }$ \\
    & OCR & $64.4_{\pm15.0}$ & $27.4_{\pm19.5}$ & $42.8_{\pm31.2}$ & $35.9_{\pm20.7}$ & $14.6_{\pm1.7 }$ & $23.2_{\pm24.0}$ & $41.9_{\pm16.9}$ & $44.6_{\pm17.5}$ \\
    & Face & $60.8_{\pm16.0}$ & $31.5_{\pm17.0}$ & $52.8_{\pm9.8 }$ & $49.3_{\pm5.6 }$ & $12.6_{\pm3.3 }$ & $32.9_{\pm14.6}$ & $43.9_{\pm4.5 }$ & $46.3_{\pm5.5 }$ \\
    & Speech & $61.8_{\pm12.8}$ & $26.6_{\pm17.6}$ & $43.8_{\pm34.5}$ & $34.2_{\pm20.6}$ & $14.4_{\pm4.8 }$ & $12.9_{\pm9.6 }$ & $40.9_{\pm16.0}$ & $43.1_{\pm16.1}$ \\
    & Concat & $68.6_{\pm8.6 }$ & $24.9_{\pm19.9}$ & $39.7_{\pm19.5}$ & $39.3_{\pm19.8}$ & $10.8_{\pm2.8 }$ & $18.3_{\pm18.1}$ & $42.2_{\pm12.3}$ & $43.2_{\pm14.2}$ \\
    \midrule
    \multicolumn{2}{c}{\textbf{MMC-PCFG}} & $\mathit{67.9}_{\pm9.8 }$ & $\mathit{52.3}_{\pm9.0 }$ & $\mathit{63.5}_{\pm8.6 }$ & $\mathit{60.7}_{\pm10.8}$ & $\mathit{34.7}_{\pm17.0}$ & $\mathit{50.4}_{\pm8.3 }$ & $\mathbf{55.0}_{\pm3.7 }$ & $\mathbf{58.9}_{\pm3.4 }$ \\
    \bottomrule
    \end{tabular}
    \caption{Performance Comparison on DiDeMo.}
    \label{tab:DiDeMo}
\end{table*}

\begin{table*}[hbt!]
\small
    \centering
    \begin{tabular}{cccccccccc}
    \toprule
    \multicolumn{2}{c}{Method} & NP & VP & PP & SBAR & ADJP & ADVP & C-F1 & S-F1 \\
    \hline
    \multicolumn{2}{c}{LBranch} & $1.7 $ & $42.8$ & $0.4 $ & $8.1 $ & $1.5 $ & $0.0 $ & $6.8 $ & $5.9 $ \\
    \multicolumn{2}{c}{RBranch} & $35.6$ & $\mathit{47.5}$ & $67.0$ & $\mathbf{88.9}$ & $\mathbf{33.9}$ & $\mathbf{65.0}$ & $35.0$ & $41.6$ \\
    \multicolumn{2}{c}{Random} & $27.2_{\pm0.3 }$ & $27.1_{\pm1.4 }$ & $29.9_{\pm0.5 }$ & $31.3_{\pm5.2 }$ & $26.9_{\pm7.7 }$ & $26.2_{\pm11.9}$ & $21.2_{\pm0.2}$ & $24.0_{\pm0.2}$ \\
    \multicolumn{2}{c}{C-PCFG} & $47.4_{\pm18.4}$ & $\mathbf{49.4}_{\pm11.9}$ & $58.0_{\pm22.6}$ & $45.7_{\pm6.0 }$ & $\mathit{27.7}_{\pm15.1}$ & $36.2_{\pm7.4 }$ & $37.8_{\pm6.7}$ & $41.4_{\pm6.6}$ \\
    \hline
    \multirow{8}{*}{\rotatebox{90}{VC-PCFG}} & ResNeXt & $46.5_{\pm13.7}$ & $40.8_{\pm9.8 }$ & $67.9_{\pm12.7}$ & $50.5_{\pm13.3}$ & $22.3_{\pm6.7 }$ & $38.8_{\pm21.3}$ & $38.2_{\pm8.3}$ & $42.8_{\pm8.4}$ \\
    & SENet & $48.3_{\pm14.4}$ & $40.7_{\pm9.2 }$ & $73.6_{\pm11.2}$ & $45.5_{\pm17.0}$ & $26.9_{\pm13.6}$ & $41.2_{\pm17.5}$ & $39.9_{\pm8.7}$ & $44.9_{\pm8.3}$ \\
    & I3D & $48.1_{\pm10.7}$ & $39.0_{\pm8.0 }$ & $\mathit{79.4}_{\pm8.4 }$ & $50.0_{\pm14.9}$ & $18.5_{\pm7.0 }$ & $41.2_{\pm4.1 }$ & $40.6_{\pm3.6}$ & $45.7_{\pm3.2}$ \\
    & R2P1D & $\mathit{52.4}_{\pm10.9}$ & $33.7_{\pm16.4}$ & $66.7_{\pm10.7}$ & $49.5_{\pm13.8}$ & $25.8_{\pm10.6}$ & $33.8_{\pm12.4}$ & $39.4_{\pm8.1}$ & $44.4_{\pm8.3}$ \\
    & S3DG & $50.4_{\pm13.1}$ & $32.6_{\pm16.3}$ & $71.7_{\pm7.5 }$ & $33.3_{\pm5.9 }$ & $30.8_{\pm17.5}$ & $40.0_{\pm7.1 }$ & $39.3_{\pm6.5}$ & $44.1_{\pm6.6}$ \\
    & Audio & $51.2_{\pm3.1 }$ & $42.0_{\pm7.2 }$ & $61.5_{\pm18.0}$ & $\mathit{51.0}_{\pm14.8}$ & $23.5_{\pm16.8}$ & $48.8_{\pm8.2 }$ & $39.2_{\pm4.7}$ & $43.3_{\pm4.9}$ \\
    & OCR & $48.6_{\pm8.1 }$ & $41.5_{\pm4.1 }$ & $65.5_{\pm17.4}$ & $39.9_{\pm4.4 }$ & $18.5_{\pm6.6 }$ & $\mathit{53.8}_{\pm14.7}$ & $38.6_{\pm5.5}$ & $43.2_{\pm5.6}$ \\
    & Concat & $50.3_{\pm10.3}$ & $42.3_{\pm2.9 }$ & $\mathbf{81.6}_{\pm8.7 }$ & $40.1_{\pm3.9 }$ & $17.7_{\pm8.2 }$ & $52.5_{\pm5.6 }$ & $\mathit{42.3}_{\pm5.7}$ & $\mathit{47.0}_{\pm5.6}$ \\
    \midrule
    \multicolumn{2}{c}{\textbf{MMC-PCFG}} & $\mathbf{62.7}_{\pm9.8 }$ & $45.3_{\pm2.8 }$ & $63.4_{\pm17.7}$ & $43.9_{\pm4.8 }$ & $26.2_{\pm7.5 }$ & $35.0_{\pm3.5 }$ & $\mathbf{44.7}_{\pm5.2}$ & $\mathbf{48.9}_{\pm5.7}$ \\
    \bottomrule
    \end{tabular}
    \caption{Performance Comparison on YouCook2.}
    \label{tab:YouCook2}
\end{table*}

\begin{table*}[hbt!]
\small
    \centering
    \begin{tabular}{cccccccccc}
    \toprule
	\multicolumn{2}{c}{Method} & NP & VP & PP & SBAR & ADJP & ADVP & C-F1 & S-F1 \\
	\hline
    \multicolumn{2}{c}{LBranch} & $34.6$ & $0.1 $ & $0.9 $ & $0.2 $ & $3.8 $ & $0.3 $ & $14.4$ & $16.8$ \\
    \multicolumn{2}{c}{RBranch} & $34.6$ & $\mathbf{90.9}$ & $67.5$ & $\mathbf{94.8}$ & $25.4$ & $54.8$ & $54.2$ & $58.6$ \\
    \multicolumn{2}{c}{Random}  & $34.6_{\pm0.1 }$ & $26.8_{\pm0.1}$ & $28.1_{\pm0.2 }$ & $24.6_{\pm0.3 }$ & $24.8_{\pm1.0}$ & $28.1_{\pm1.4}$ & $27.2_{\pm0.1}$ & $30.5_{\pm0.1}$ \\
    \multicolumn{2}{c}{C-PCFG}  & $46.6_{\pm3.2 }$ & $61.1_{\pm3.3}$ & $72.5_{\pm8.3 }$ & $63.7_{\pm4.0 }$ & $33.1_{\pm7.1}$ & $67.1_{\pm4.7}$ & $50.7_{\pm3.2}$ & $55.0_{\pm3.2}$ \\
    \hline
    \multirow{11}{*}{\rotatebox{90}{VC-PCFG}} & ResNeXt & $48.6_{\pm3.0 }$ & $59.0_{\pm6.0}$ & $72.0_{\pm3.6 }$ & $62.1_{\pm5.2 }$ & $32.6_{\pm2.5}$ & $70.4_{\pm6.4}$ & $50.7_{\pm1.7}$ & $54.9_{\pm2.2}$ \\
    & SENet & $49.0_{\pm4.4 }$ & $63.5_{\pm6.4}$ & $71.7_{\pm4.8 }$ & $60.9_{\pm10.6}$ & $34.0_{\pm6.4}$ & $\mathit{74.1}_{\pm1.9}$ & $52.2_{\pm1.2}$ & $56.0_{\pm1.6}$ \\
    & I3D & $\mathbf{53.9}_{\pm10.5}$ & $63.2_{\pm9.1}$ & $73.7_{\pm2.9 }$ & $65.3_{\pm9.1 }$ & $\mathbf{35.0}_{\pm6.8}$ & $73.8_{\pm4.1}$ & $54.5_{\pm1.6}$ & $\mathit{59.1}_{\pm1.7}$ \\
    & R2P1D & $52.8_{\pm3.6 }$ & $63.3_{\pm4.6}$ & $73.1_{\pm10.1}$ & $66.9_{\pm2.0 }$ & $34.0_{\pm2.2}$ & $72.5_{\pm4.2}$ & $54.0_{\pm2.5}$ & $58.0_{\pm2.3}$ \\
    & S3DG & $48.2_{\pm4.4 }$ & $60.4_{\pm3.9}$ & $71.4_{\pm6.4 }$ & $58.1_{\pm8.2 }$ & $25.3_{\pm2.2}$ & $61.8_{\pm8.4}$ & $50.7_{\pm3.2}$ & $54.7_{\pm2.9}$ \\
    & Scene & $50.7_{\pm1.6 }$ & $65.0_{\pm4.7}$ & $\mathbf{78.6}_{\pm3.6 }$ & $\mathit{67.3}_{\pm3.9 }$ & $\mathit{34.5}_{\pm4.6}$ & $71.7_{\pm1.8}$ & $\mathit{54.6}_{\pm1.5}$ & $58.4_{\pm1.3}$ \\
    & Audio & $50.0_{\pm1.1 }$ & $63.7_{\pm6.1}$ & $72.7_{\pm3.0 }$ & $61.9_{\pm6.5 }$ & $\mathit{34.5}_{\pm2.3}$ & $68.0_{\pm5.9}$ & $52.8_{\pm1.3}$ & $56.7_{\pm1.4}$ \\
    & OCR & $48.3_{\pm8.3 }$ & $57.1_{\pm4.6}$ & $76.9_{\pm0.6 }$ & $60.7_{\pm4.9 }$ & $33.9_{\pm8.3}$ & $72.1_{\pm4.4}$ & $51.0_{\pm3.0}$ & $55.5_{\pm3.0}$ \\
    & Face & $46.5_{\pm6.8 }$ & $61.3_{\pm3.6}$ & $71.5_{\pm7.1 }$ & $60.8_{\pm11.0}$ & $30.9_{\pm3.4}$ & $68.4_{\pm6.0}$ & $50.5_{\pm2.6}$ & $54.5_{\pm2.6}$ \\
    & Speech & $48.5_{\pm7.6 }$ & $60.7_{\pm3.5}$ & $74.5_{\pm5.7 }$ & $62.6_{\pm6.2 }$ & $27.3_{\pm1.8}$ & $74.0_{\pm3.1}$ & $51.7_{\pm2.6}$ & $56.2_{\pm2.5}$ \\
    & Concat & $43.6_{\pm6.0 }$ & $64.7_{\pm3.0}$ & $68.5_{\pm8.0 }$ & $63.8_{\pm3.8 }$ & $32.0_{\pm5.5}$ & $70.4_{\pm5.9}$ & $49.8_{\pm4.1}$ & $54.2_{\pm4.0}$ \\
    \midrule
    \multicolumn{2}{c}{\textbf{MMC-PCFG}} & $\mathit{52.3}_{\pm5.1 }$ & $\mathit{68.1}_{\pm2.9}$ & $\mathit{78.2}_{\pm1.9 }$ & $65.8_{\pm2.4 }$ & $32.0_{\pm2.0}$ & $\mathbf{74.7}_{\pm2.3}$ & $\mathbf{56.0}_{\pm1.4}$ & $\mathbf{60.0}_{\pm1.2}$ \\
    \bottomrule
    \end{tabular}
    \caption{Performance Comparison on MSRVTT.}
    \label{tab:MSRVTT}
\end{table*}
\clearpage

\end{document}